\documentclass{article}
\usepackage{graphicx} 
\usepackage[a4paper,top=3cm,bottom=2cm,left=2.3cm,right=2.3cm,marginparwidth=1.75cm]{geometry}
\usepackage{graphicx}
\usepackage{verbatim}
\usepackage[hidelinks]{hyperref}       
\usepackage{url}            
\usepackage{booktabs}       
\usepackage{amsfonts}       
\usepackage{nicefrac}       
\usepackage{microtype}      
\usepackage{amsmath}
\usepackage{multirow}
\usepackage{amsthm}
\usepackage{amssymb}
\newtheorem{theorem}{Theorem}[section]

\newtheorem{definition}{Definition}[section]

\usepackage{subfigure}
\usepackage{times}
\usepackage{color}
\newcommand{\bL}{\boldsymbol{L}}
\newcommand{\bx}{\boldsymbol{x}}
\newcommand{\bX}{\boldsymbol{X}}

\newcommand{\bk}{\boldsymbol{k}}
\newcommand{\bG}{\boldsymbol{G}}
\newcommand{\bw}{\boldsymbol{w}}
\newcommand{\bdf}{\boldsymbol{f}}

\newcommand{\bS}{\boldsymbol{S}}

\newcommand{\bI}{\boldsymbol{I}}
\newcommand{\bA}{\boldsymbol{A}}
\newcommand{\bB}{\boldsymbol{B}}

\newcommand{\bSigma}{\boldsymbol{\Sigma}}

\usepackage{algorithm}
\usepackage{algpseudocode}

\title{Score-fPINN: Fractional Score-Based Physics-Informed Neural Networks for High-Dimensional Fokker-Planck-L\'evy Equations}
\author{Zheyuan Hu\thanks{Department of Computer Science, National University of Singapore, Singapore, 119077 (\href{mailto:e0792494@u.nus.edu}{e0792494@u.nus.edu},\href{mailto:kenji@nus.edu.sg}{kenji@nus.edu.sg})} \and Zhongqiang Zhang\thanks{Department of Mathematical Sciences, Worcester Polytechnic Institute, Worcester, MA 01609 USA (\href{mailto:zzhang7@wpi.edu}{zzhang7@wpi.edu})}\and  George Em Karniadakis\thanks{Division of Applied Mathematics, Brown University, Providence, RI 02912, USA (\href{mailto:george\_karniadakis@brown.edu}{george\_karniadakis@brown.edu})} \ \thanks{Advanced Computing, Mathematics and Data Division, Pacific Northwest National Laboratory, Richland, WA, United States} \and Kenji Kawaguchi\footnotemark[1]}
\date{}

\begin{document}

\maketitle

\begin{abstract}
We introduce an innovative approach for solving high-dimensional Fokker-Planck-Lévy (FPL) equations in modeling non-Brownian processes across disciplines such as physics, finance, and ecology. 
We utilize a fractional score function and 
Physical-informed neural networks (PINN) to lift the curse of dimensionality (CoD) and 
alleviate numerical overflow from exponentially decaying solutions with dimensions. 
The introduction of a fractional score function
allows us to transform the FPL equation into a 
second-order partial differential equation without fractional Laplacian and thus can be readily solved with standard physics-informed neural networks (PINNs).
We propose two methods to obtain a fractional score function: fractional score matching (FSM) and score-fPINN for fitting the fractional score function. While FSM is more cost-effective, it relies on known conditional distributions. On the other hand, score-fPINN is independent of specific stochastic differential equations (SDEs) but requires evaluating the PINN model's derivatives, which may be more costly.
We conduct our experiments on various SDEs and 
demonstrate numerical stability and effectiveness of our method in dealing with high-dimensional problems, marking a significant advancement in addressing the CoD in FPL equations.

\end{abstract}


\section{Introduction}
The Fokker-Planck-L\'evy (FPL) equation, also known as the fractional Fokker-Planck equation, is a generalization of the traditional Fokker-Planck (FP) equation to incorporate L\'evy processes, particularly those involving jumps or heavy-tailed distributions. The FPL equation is used in fields like physics for anomalous diffusion in complex systems, finance for pricing derivatives when the underlying asset exhibits jumps or heavy tails, and ecology for animal movement patterns that involve sudden, long-range moves.
The classic Fokker-Planck equation describes the time evolution of the probability density function of the velocity of a particle under the influence of forces and Gaussian white noises. 
However, many physical and economic phenomena exhibit jumps and heavy tails, which are not adequately described by Gaussian processes. L\'evy processes, which include a broader class of stochastic processes characterized by stable distributions and jumps, offer a more appropriate mathematical framework for such scenarios. The FPL equation extends the traditional Fokker-Planck equation by incorporating fractional derivatives, which can model these non-local, jump-like dynamics.

Despite its importance, obtaining numerical solutions to the FPL equation is still challenging due to the non-locality introduced by the fractional derivative, which requires special numerical schemes that can handle integral terms effectively and the need for handling both the small-scale behavior driven by the diffusion term as well as the large discrete changes introduced by the jump term.

The higher-dimensional FPL equations of interest in this paper pose more significant challenges owing to 
the curse of dimensionality (CoD), where traditional grid-based methods fail due to the exponential increase in computational requirements with the dimensionality of the PDE, rendering them unrealistic. Consider another branch of the traditional method, namely Monte Carlo simulation, which can tackle the CoD in certain PDEs. Though Monte Carlo methods can solve the FP equation with the Feynman–Kac formula and the corresponding stochastic differential equation (SDE), they solve the problem at one point and are also expensive when the solution at a large region is desired. 

Physics-informed neural networks \cite{raissi2019physics} have become popular in solving high-dimensional PDEs and tackling the CoD thanks to neural networks' strong universal approximation property \cite{barron1993universal}, generalization capacities \cite{hu2021extended}, robust optimization \cite{kawaguchi2021theory}, and PINN's meshless plus grid-free training.
Multiple methods \cite{hu2023hutchinson,hu2023tackling} recently have proposed to scale up and speed up PINNs to very high dimensions using random sampling. 
Although PINN offers the possibility of addressing the CoD in certain cases, in FPL equations, PINN accuracy is limited to moderately high dimensions (e.g., less than ten dimensions \cite{chen2021solving}) for computing probability density functions (PDFs) of interest in the FPL equations. 
They also exhibit significant numerical errors at higher dimensions, rendering them impractical. Specifically, FPL equations model PDFs and the most common Gaussian PDFs associated with Brownian motion exhibit exponential decay in numerical values as dimensionality increases. This phenomenon easily surpasses the numerical precision of computer simulations, leading to significant errors in PINN numerical solvers. 
The heavy-tailed nature of Levy noise results in rare events with extremely small PDF values, further amplifying the numerical errors of vanilla PINNs.
Furthermore, the above high-dimensional PINNs \cite{hu2023hutchinson,hu2023tackling} works are based on integer-order derivatives and employ automatic differentiation to compute integer-order derivatives.
%
However, fractional-order derivatives are not yet covered by automatic differentiation libraries. 
Approximating the \textit{high-dimensional} fractional Laplacian is still an open problem.
Therefore, tackling high-dimensional FPL equations remains a challenging problem, whether for traditional methods or emerging techniques like PINNs.

To this end, we propose utilizing a fractional score-based SDE/FPL equation solver to fit the fractional score function in the SDE, which broadens the definition of the score function used in the FP equation \cite{hyvarinen2005estimation,song2021scorebased,hu2024score}.
We demonstrate its numerical stability and the fundamental role of fractional score in solving the FPL equation's SDE, indicating its capability to accurately infer the log-likelihood (LL) and PDF of interest. Concretely, with the obtained fractional score, we can eliminate the fractional Laplacian in the FPL equation, transforming it into an FP equation that can be solved using standard PINNs \cite{raissi2019physics,hu2023tackling,hu2023hutchinson}. The second-order FP equation enables subsequent fitting of LL and PDF via standard PINNs afterward.

We introduce two methods for fitting the crucial fractional score function: fractional score matching (FSM) \cite{yoon2023scorebased} and score-fPINN.
FSM relies on the conditional distribution modeled by the SDE and it minimizes the mean squared error between the fractional score model and the true conditional fractional score along SDE trajectories. One can prove that this is equivalent to minimizing the mean squared error between the fractional score model and the true fractional score \cite{yoon2023scorebased}. 
When the conditional distribution modeled by the SDE is unknown, we use Score-fPINN which is independent of the SDE form.  Specifically, Score-fPINN first utilizes Sliced Score Matching (SSM) \cite{song2019sliced} to obtain the conventional score function, which is then inserted into the FPL equation to simplify it. Then, the only unknown will be the fractional score function, which can be obtained by enforcing the PINN's PDE loss.
Once the score function is obtained through one of the above methods, we simplify the FPL equation to a second-order FP equation, allowing easy calculation of the LL using standard PINN methods. In other words, our fractional score-based SDE/FPL equation solver consists of two stages. The first stage involves FSM or Score-fPINN to obtain the fractional score function. The second stage requires using the obtained fractional score function to solve LL or PDF via a second-order FP equation, which is obtained from plugging the fractional score function into the original FPL equation.
In comparison between these two methods, FSM is more concise and efficient but requires the known conditional distribution from the SDE.
Score-fPINN employs PINN loss and thus necessitating the calculation of derivatives of the neural network model with respect to the input. Thus, Score-PINN is more expensive while score-fPINN applies to a broader range of SDE types.

We evaluate the fractional score-based SDE/FPL equation solver on different SDEs.
We test the basic case, namely, an anisotropic SDE with both Brownian and L\'evy noise, and then test more challenging problems by complicating its diffusion and drift coefficients. We also vary the initial distribution to assess the capability of our framework to fit different distributions.
Experimental results demonstrate the stability of the fractional score-based SDE solver in various experimental settings.
The proposed methods are sublinear in speed, and their performance remains stable across dimensions, which demonstrates the ability of the fractional score-based SDE solver to overcome CoD in FPL equations.

To the best of our knowledge, we introduce the concept of fractional score function and the Score-fPINN in solving high-dimensional FPL equations to the scientific machine learning community for the first time. 

\section{Related Work}
\subsection{PINN for FP and FPL equations}
FP and FPL equations are prevalent in statistical mechanics, and their high dimensionality poses significant challenges to traditional analytical methods such as finite difference \cite{deng2009finite,sepehrian2015numerical}. 
In contrast, machine learning techniques, particularly physics-informed neural networks (PINNs), offer a promising mesh-free solution capable of addressing the CoD and integrating observational data smoothly. 
Research by Chen et al. \cite{chen2021solving} involved using PINNs to address both forward and inverse issues associated with Fokker-Planck-L\'evy equations. Similarly, Zhang et al. \cite{zhang2022solving} applied deep KD-tree methods to tackle FP equations in scenarios with sparse data. Zhai et al. \cite{zhai2022deep} and Wang et al. \cite{wang2024tensor} utilized deep learning for solving steady-state FP equations, while Lu et al. \cite{lu2022learning} focused on learning high-dimensional multivariate probability densities modeled by FP equations using normalizing flows. Furthermore, Feng et al. \cite{feng2021solving} and Guo et al. \cite{guo2022normalizing} both employed normalization flow techniques for FP equations, and Tang et al. \cite{tang2022adaptive} introduced an adaptive deep density approximation method based on normalizing flows for steady-state FP equations. Hu et al. \cite{hu2024score} introduced a score-based SDE solver for FP equations, and herein we introduce their methodology to the fractional score and FPL equation settings.

\subsection{High-Dimensional PDE Solvers} Numerous techniques have been developed to overcome the curse of dimensionality (COD) in solving high-dimensional partial differential equations (PDEs): physics-informed neural networks (PINNs) \cite{raissi2019physics,sirignano2018dgm}, backward stochastic differential equations (BSDE) \cite{han2018solving}, and the Multilevel Picard method \cite{beck2020overcoming}. Specifically, using the PINN framework, He et al. \cite{he2023learning} introduced the randomized smoothing PINN (RS-PINN), which leverages Monte Carlo simulations and Stein's identity to evaluate derivatives, thus bypassing expensive automatic differentiation techniques. Subsequently, Zhao et al. \cite{zhao2023tensor} suggested replacing Monte Carlo simulations with sparse quadratures to lower variance, while Hu et al. \cite{hu2023bias} investigated the bias-variance dilemma in RS-PINN. Stochastic dimension gradient descent (SDGD) \cite{hu2023tackling} reduces memory usage and hastens convergence by sampling subsets of dimensions for gradient descent when training PINNs. Meanwhile, Hutchinson trace estimation (HTE) \cite{hu2023hutchinson} offers an alternative to the high-dimensional Hessian by a Hessian vector product based on HTE in the PINN loss function to speed up the process.
Regarding BSDE, the backward stochastic differential equations method \cite{han2018solving} and deep splitting approach \cite{Beck2019DeepSM} integrate deep learning with traditional techniques for solving parabolic PDEs, allowing for the modeling of unknown functions within these established frameworks.
Lastly, the multilevel Picard method \cite{beck2020overcoming,hutzenthaler2021multilevel} addresses nonlinear parabolic PDEs under specific regularity conditions for convergence.

\subsection{Fractional PDE Solvers}
Pang et al. \cite{pang2019fpinns} propose fractional PINN (fPINN), adopting neural networks as solution surrogates and discretizing the fractional derivative for supervision.
Guo et al. \cite{guo2022monte} propose Monte Carlo fPINN (MC-fPINN) estimating Caputo-type time-fractional derivatives and fractional Laplacian in the hyper-singular integral representation using Monte Carlo from Beta distributions.
Firoozsalari et al. \cite{firoozsalari2023deepfdenet} consider Gaussian quadrature to compute the numerical integral related to fractional derivative.
Leonenko and Podlubny \cite{leonenko2022monte} propose a Monte Carlo-based estimator for the Grünwald–Letnikov fractional derivative.
While these methods, especially MC-fPINN \cite{guo2022monte}, effectively approximate the high-dimensional fractional Laplacian in FPL equations, they are still constrained by the numerical instability caused by the exceedingly small values of the PDF modeled by the FPL equation.

\subsection{Score-Based Generative Models}
Song et al. \cite{song2021scorebased} highlighted the relationship between diffusion generative models and SDEs. These SDEs inject noise into data sets, such as images and texts, converting them to a pure Gaussian state. The reverse process of the SDE then removes the noise to restore the original data distribution. Central to this mechanism is the derivation of the score function, which is the gradient of the log-likelihood of a distribution, an approach known as score matching (SM). Various techniques for score matching have been developed. For instance, Song et al. \cite{song2021scorebased} developed methods for matching the conditional score function, a technique that is mathematically on par with direct score matching \cite{hyvarinen2005estimation}. Moreover, Song et al. \cite{song2019sliced} introduced sliced score matching (SSM), achieving objectives similar to direct score matching but without necessitating knowledge of the underlying distribution. Finite difference score matching \cite{pang2020efficient} further lessens the computational burden associated with sliced score matching by sidestepping the intensive computation of gradients in the score function in SSM. Additionally, Lai et al. \cite{lai2022regularizing} explored the partial differential equation that governs the score function in FP equations, suggesting the use of physics-informed neural networks (PINNs) \cite{raissi2019physics} to streamline the score-matching optimization process.
Boffi and Vanden-Eijnden \cite{boffi2023deep,boffi2023probability} adopt score-based solvers to time-dependent and time-independent Fokker-Planck equations.
\cite{yoon2023scorebased} extend the score matching based on SDE with Brownian motion to the L\'evy process, transforming the data distribution to stable distributions, and further propose the corresponding fractional score matching, which is used to reverse the L\'evy process for data generation from random L\'evy noise.

\begin{table}[!ht]
\centering
\caption{List of Abbreviations}
\begin{tabular}{|c|c|}
\hline
\textbf{Abbreviation} & \textbf{Explanation} \\
\hline
CoD & Curse-of-Dimensionality \\\hline
PDE & Partial Differential Equation \\\hline
ODE & Ordinary Differential Equation \\\hline
SDE & Stochastic Differential Equation \\\hline
PDF & Probability Density Function \\\hline
LL & Logarithm Likelihood \\\hline
PINN & Physics-Informed Neural Network\\\hline
fPINN & Fractional Physics-Informed Neural Network\\\hline
HTE & Hutchinson Trace Estimation\\\hline
HJB & Hamilton-Jacobi-Bellman\\\hline
FP & Fokker-Planck\\\hline
FPL & Fokker-Planck-L\'evy\\\hline
SM & Score Matching\\\hline
SSM & Sliced Score Matching\\\hline
FSM & Fractional Score Matching\\\hline
OU & Ornstein–Uhlenbeck\\\hline
AI & Artificial Intelligence\\\hline
\end{tabular}
\label{tab:scorefpinn_abbreviations}
\end{table}

\begin{table}[!ht]
\centering
\caption{List of Notations}
\begin{tabular}{|c|c|}
\hline
\textbf{Notation} & \textbf{Explanation} \\
\hline
$\mathcal{S}\alpha\mathcal{S}^d(\gamma)$ & $d$-dimensional $\alpha$-stable distribution with parameter $\gamma$\\\hline 
$p_t(\bx)$ & Probability density function (PDF) concerning time $t$ and input $\bx$\\\hline
$\bdf(\bx, t)$ & Drift coefficient of the SDE \\\hline
$\bG(\bx, t)$ & Diffusion coefficient of the SDE \\\hline
$\sigma(t)$ & Coefficient for the L\'evy noise in the SDE \\\hline
$\bA^\alpha(\bx, t)$ & $\boldsymbol{A}^\alpha(\bx, t) := \bdf(\bx, t) - \frac{1}{2}\nabla \cdot \left[\bG(\bx, t)\bG(\bx, t)^\mathrm{T}\right]- \sigma(t)\bS_t^{(\alpha)}(\bx)$\\\hline
$\bw_t$ & Brownian motion \\\hline
$\bL_t^\alpha$ & L\'evy process \\\hline
$q_t(\bx)$ & Logarithm likelihood (LL) of $p_t(\bx)$, i.e., $q_t(\bx) = \log p_t(\bx)$ \\\hline
$q_t(\bx; \phi)$ & PINN model parameterized by $\phi$ to model LL\\\hline
$\bS_t^{(\alpha)}(\bx)$ & Fractional score function defined as $\bS_t^{(\alpha)}(\bx) = \frac{(-\Delta)^{\frac{\alpha-2}{2}}\nabla p_t(\bx)}{p_t(\bx)}$\\\hline
$\bS_t^{(\alpha)}(\bx;\theta)$ & Fractional score function PINN model parameterized by $\theta$, i.e., Score-fPINN\\\hline
\end{tabular}
\label{tab:scorefpinn_notations}
\end{table}

\section{Proposed Method}
This section presents the methodology of employing a fractional score-based model to address SDE forward problems with Brownian and L\'evy noises. We list abbreviations and notations in Tables \ref{tab:scorefpinn_abbreviations} and \ref{tab:scorefpinn_notations}.

\subsection{Problem Definition and Background}
\subsubsection{Stable Distribution}
\begin{figure}[htbp]
    \centering
    \includegraphics[scale=0.32]{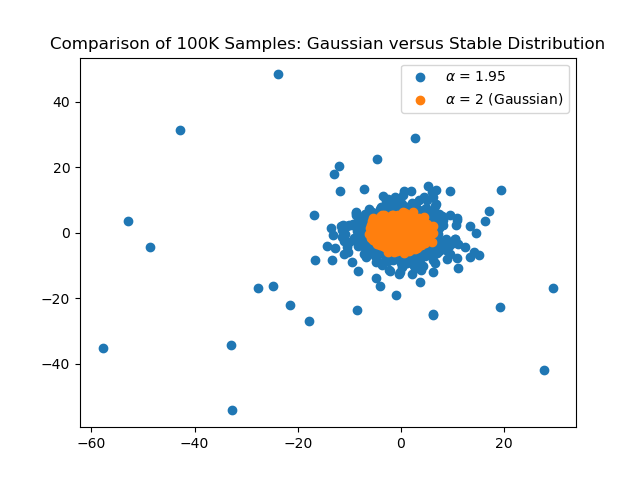}
    \includegraphics[scale=0.32]{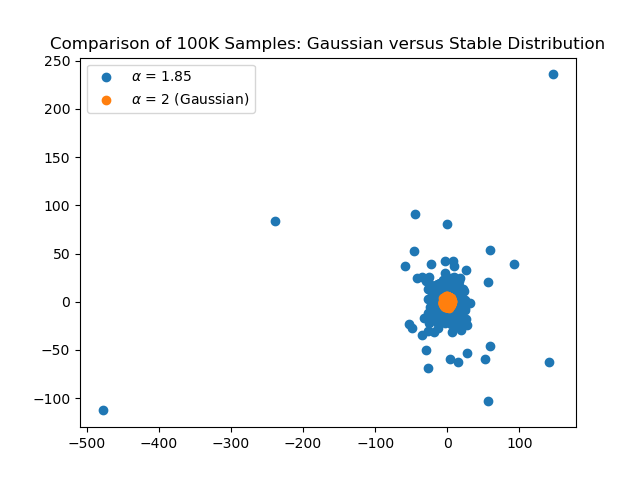}
    \includegraphics[scale=0.32]{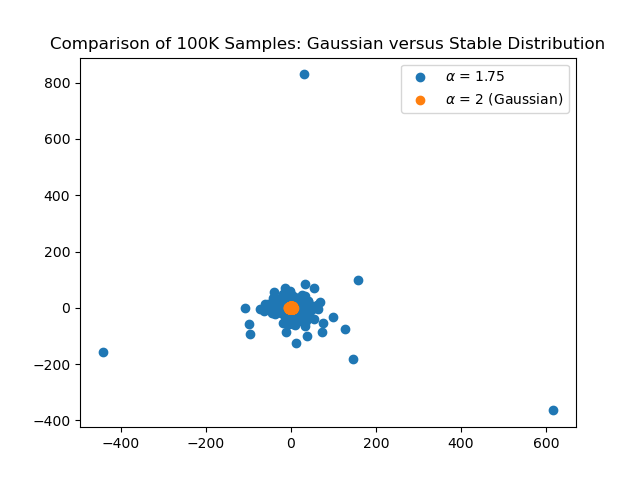}
    \caption{We generate 100,000 random samples from stable distributions with various $\alpha$ in 2D for visualization. The long tail of $\alpha < 2$ stable distributions exhibit a much larger support than the short tail of Gaussian with $\alpha = 2$. Rare events far from the original point are also common for stable distributions with $\alpha < 2$.}
    \label{fig:compare_gaussian_levy}
\end{figure}

\begin{definition} ($\alpha$-stable L\'evy distribution)
If a random variable $X \sim \mathcal{S}\alpha\mathcal{S}^d(\gamma) \in \mathbb{R}^d$, then its characteristic function $\mathbb{E}[\exp(i\langle \bk, X\rangle)] = \exp(-\gamma^\alpha\Vert\bk\Vert^\alpha)$.
\end{definition}
Here are some examples of analytical L\'evy distributions:
\begin{itemize}
\item If $\alpha=2$, then $\mathcal{S}\alpha\mathcal{S}^d(\gamma) = \mathcal{N}(0, 2\gamma^2 I)$ is Gaussian.
\item If $\alpha=1$, then $\mathcal{S}\alpha\mathcal{S}^d(\gamma)$ is the Cauchy distribution with the PDF \cite{kotz2004multivariate,lee2014clarification}:
\begin{equation}
p(\bx; \gamma) = \frac{\Gamma\left(\frac{1 + d}{2}\right)}{\Gamma\left(\frac{1}{2}\right)(\gamma^2\pi)^{\frac{d}{2}}\left[(1 + \frac{\Vert\bx\Vert^2}{ \gamma^2})\right]^{\frac{d+1}{2}}}.
\end{equation}
Its characteristic function is $\phi(\bk) = \exp(- \gamma \Vert \bk \Vert)$.
\item If $\alpha=1.5$, then $\mathcal{S}\alpha\mathcal{S}^d(\gamma)$ is the Holtsmark distribution whose PDF can be expressed via hypergeometric functions.
\item For stable distribution with other $\alpha$, the PDF does not have a closed-form expression.
\end{itemize}
When $\alpha = 2$, the Gaussian distribution has an exponentially short tail, e.g., for the 1D Gaussian $X \sim \mathcal{N}(0, \sigma^2)$, its tail bound is 
$
\mathbb{P}(X \geq r) \leq \exp\left(-\frac{r^2}{2\sigma^2}\right).
$
Modeling stable distributions is difficult due to its long tail property when $\alpha<2$. 
When $\alpha < 2$, let $\boldsymbol{X}$ follow the $d$-dimensional $\alpha$-stable distribution and it has a long tail, i.e.,
$
\mathbb{P}(\Vert \boldsymbol{X} \Vert_2 \geq r) \sim r^{-\alpha}.
$
Thus, rare events are relatively common for stable distributions compared with the short-tailed Gaussian, as shown in Figure \ref{fig:compare_gaussian_levy}.
Due to the long tail, for $\alpha < 2$, the variance of $\alpha$-stable distribution is infinite.
We mainly focus on $1 < \alpha < 2$.

\subsubsection{L\'evy Process}
Furthermore, a L\'evy process $\bL_t$ is a stochastic process that generalizes Brownian motion by allowing non-Gaussian increments.
\begin{definition} (L\'evy Process) An $\mathbb{R}^d$-valued stochastic process $\bL_t$ is a L\'evy process if
\begin{enumerate}
\item $\bL_0 = 0$ almost surely.
\item $\bL_t$ has independent increments.
\item $\bL_t$ has stationary increments.
\item The sample paths of the process are stochastically continuous.
\end{enumerate}
\end{definition}
\noindent 
Under the $\alpha$-stable L\'evy distribution, the isotropic $\alpha$-stable L\'evy process $\bL_t^\alpha$ under consideration satisfies that for all $s < t$, $\bL_t^\alpha  - \bL_s^\alpha =\bL_{t-s}^\alpha \sim \mathcal{S}\alpha\mathcal{S}^d((t - s)^{1/\alpha})$ in distribution.

The PDF takes extremely small values in most regions and decays exponentially with dimensions and thus it causes numerical error to conventional PINNs \cite{raissi2019physics}.
Moreover, The heavy-tail property of $\alpha$-stable L\'evy noise leads to the huge support of the PDFs, making domain truncation impractical.
%

\subsubsection{Fractional SDE and Fokker-Planck-L\'evy Equation}

Then, we consider SDEs with Brownian and $\alpha$-stable L\'evy noises:
\begin{equation}\label{eq:sde_levy}
d\bX = \bdf(\bX, t)dt + \bG(\bX, t)d\bw_t + \sigma(t)d\bL_t^\alpha,
\end{equation}
where $\bL_t^\alpha$ is the $\alpha$-stable L\'evy  process in $\mathbb{R}^d$ and $\bX \in \mathbb{R}^d$ is the state variable, $t \in [0, T]$ where $T$ is the terminal time.
Here all the coefficients are known: $\bdf(\bx, t): \mathbb{R}^d \times \mathbb{R} \rightarrow \mathbb{R}^d$, $\bG(\bx, t): \mathbb{R}^d \times \mathbb{R} \rightarrow \mathbb{R}^{d\times d}$, and $\sigma(t): \mathbb{R} \rightarrow \mathbb{R}$.
We want to solve the forward problem, i.e., given the initial distribution $p_0(\bx)$ at $t=0$ and the SDE coefficients, we solve the evolution of the distribution of $\bx$ on $[0,T]$.
The PDF for $\mathbf{X}$ satisfies the following fractional FPL equation:
\begin{equation}
\partial_t p_t(\bx) = -\sum_{i=1}^d \frac{\partial}{\partial \bx_i} \left[\bdf_i(\bx, t)p_t(\bx)\right]+\frac{1}{2}\sum_{i,j=1}^d\frac{\partial^2}{\partial \bx_i \partial \bx_j} \left[\sum_{k=1}^d\bG_{ik}(\bx, t)\bG_{jk}(\bx, t)p_t(\bx)\right] - \sigma(t)(-\Delta)^{\frac{\alpha}{2}}p_t(\bx),
\end{equation}
where the fractional Laplacian $(-\Delta)^{\frac{\alpha}{2}}$ is formally defined as follows using the Fourier transform \cite{LISCHKE2020109009,Stinga+2019+235+266}.
\begin{definition}\label{defn:fractional-laplacian-fourier}
Denote the Fourier transform of the function $f$ as $\mathcal{F}\{f\}(\bk) = \int_{\mathbb{R}^d}\exp\left(i\langle \bx, \bk \rangle\right)f(\bx)d\bx$. 
The Fractional Laplacian for $\alpha \in (0, 2)$ is defined as
\begin{align}
(-\Delta)^{\frac{\alpha}{2}} f(\bx)=\frac{1}{(2\pi)^d} \int_{\mathbb{R}^d}\Vert \bk \Vert^\alpha\exp\left(-i\langle \bx, \bk \rangle\right)\mathcal{F}\{f\}(\bk)d\bk.
\end{align}
\end{definition}
\noindent
Direct computation of the fractional Laplacian to solve the fractional FPL equation corresponding to the L\'evy process suffers from the curse of dimensionality due to the high-dimensional integral for traditional numerical schemes \cite{gao2016fokker}.

\subsection{Fractional Score Function}
We introduce the fractional score function $\bS_t^{(\alpha)}(\bx)$ that generalizes the conventional score function \cite{hyvarinen2005estimation,song2021scorebased} to enable the solution of the LL and circumvent CoD. 
Specifically, the fraction score functions for an underlying PDF $p_t(\bx)$ are defined as follows \cite{yoon2023scorebased}:
\begin{equation}
\bS_t^{(\alpha)}(\bx) = \frac{(-\Delta)^{\frac{\alpha-2}{2}}\nabla p_t(\bx)}{p_t(\bx)}.
\end{equation}
If $\alpha = 2$, it becomes a vanilla score function, and the L\'evy process is Brownian, and $\bS_t^{(2)}(\bx) = \nabla \log p_t(\bx) = \frac{\nabla p_t(\bx)}{p_t(\bx)}$.  The vanilla score function of multivariate Gaussian $\mathcal{N}(\boldsymbol{\mu}, \boldsymbol{\Sigma})$ is
$
-\boldsymbol{\Sigma}^{-1} (\bx - \boldsymbol{\mu}).
$
It converts the inseparable multivariate Gaussian PDF to a linear score function. Similarly, the fractional score of any L\'evy distribution is also a linear function.
\begin{theorem} (Yoon et al. \cite{yoon2023scorebased})
If a random variable $\bx \sim \mathcal{S}\alpha\mathcal{S}^d(\gamma) \in \mathbb{R}^d$ and its PDF is denote $p(\bx)$, then its fractional score is
$
\bS^{(\alpha)}(\bx) = \frac{(-\Delta)^{\frac{\alpha-2}{2}}\nabla p(\bx)}{p(\bx)} = -\frac{\bx}{\alpha\gamma^\alpha}.
$
\end{theorem}
%
Compared with PDF, whose value decays exponentially with SDE dimensionality, the fractional score function's scale is invariant with dimension, making it numerically stable and preventing the PINN training from numerical overflow.

Another pivotal advantage of the fractional score is that
the fractional Laplacian can be simplified 
as the negative of the divergence operator.
By Definition \ref{defn:fractional-laplacian-fourier} of the fractional Laplacian and by the Fourier transform (see also Lemma C.1 in \cite{yoon2023scorebased}), 
\begin{align}
(-\Delta)^{\frac{\alpha}{2}} p_t(\bx) = -\nabla \cdot\left((-\Delta)^{\frac{\alpha-2}{2}}\nabla p_t(\bx)\right) =  -\nabla \cdot\left( p_t(\bx) S_t^{(\alpha)}(\bx)\right).
\end{align}
Then,  we can rewrite the FPL equation as follows 
\begin{align}
\partial_t p_t(\bx) &= -\nabla_{\bx} \cdot \left[\left(\bdf(\bx, t) - {\sigma(t)^\alpha \bS^{(\alpha)}_t(\bx)}\right)p_t(\bx)\right] - \frac{1}{2}\sum_{i,j=1}^d\frac{\partial^2}{\partial \bx_i \partial \bx_j} \left[\sum_{k=1}^d\bG_{ik}(\bx, t)\bG_{jk}(\bx, t)p_t(\bx)\right].
\end{align}
Taking $\nabla$ at both sides and dividing by $p_t(\bx)$ gives 
a second-order PDE (LL-PDE) for the LL $q = q_t(\bx) = \log p_t(\bx)$:   
\begin{align}\label{eq:LL_levy_pde}
\partial_t q &= \frac{1}{2}\nabla_{\bx} \cdot(\bG\bG^\mathrm{T}\nabla_{\bx} q) + \frac{1}{2}\Vert\bG^\mathrm{T}\nabla_{\bx} q\Vert^2 - \langle \boldsymbol{A}^\alpha, \nabla_{\bx} q\rangle - \nabla_{\bx} \cdot \boldsymbol{A}^\alpha := \mathcal{L}_{\text{LL-PDE}}\left[q, \bS_t^{(\alpha)}\right],
\end{align}
where
\begin{align}\label{eq:scorefpinn_A_alpha}
\boldsymbol{A}^\alpha(\bx, t) = \bdf(\bx, t) - \frac{1}{2}\nabla \cdot \left[\bG(\bx, t)\bG(\bx, t)^\mathrm{T}\right]- \sigma(t)\bS_t^{(\alpha)}(\bx).
\end{align}
Here, we define the LL-PDE operator $\mathcal{L}_{\text{LL-PDE}}\left[q, \bS_t^{(\alpha)}\right]$ since $\bdf, \bG, \sigma(t)$ are known and only the LL $q$ and the fractional score function $\bS^{(\alpha)}_t(\bx)$ will be the input variables. 
Hence, it is evident that once we acquire the fractional score, solving the LL can be realized by using PINNs for this second-order PDE. It is then crucial to obtain the fractional score and
we will introduce two methods to obtain it: fractional score matching and score fractional PINN (Score-fPINN).
\subsection{Fractional Score Matching (FSM)}
Fractional Score Matching (FSM) aims to parameterize $\bS_t^{(\alpha)}(\bx; \theta)$ to approximate the fractional score function $\bS_t^{(\alpha)}(\bx)$, and minimize their $L_2$ distance:
\begin{align}
L^\alpha_{\text{Oracle-SM}}(\theta) &= \mathbb{E}_{t \sim \text{Unif}[0, T]}\mathbb{E}_{\bx \sim p_t(\bx)} \left[\left\|\bS_t^{(\alpha)}(\bx; \theta) - \bS^{(\alpha)}_t(\bx) \right\|^2\right].
\end{align}
However, the exact fractional score $\bS^{(\alpha)}_t(\bx)$ is given by the exact SDE solution, which is unknown.
So, computing the objective function $L^\alpha_{\text{Oracle-SM}}(\theta)$ is unrealistic.
Yoon et al. \cite{yoon2023scorebased} demonstrate that minimizing $L^\alpha_{\text{Oracle-SM}}(\theta)$ is equivalent to minimizing conditional score matching loss function.
\begin{align}\label{eq:loss_fsm}
L^\alpha_{\text{Cond-SM}}(\theta) &= \mathbb{E}_{t \sim \text{Unif}[0, T]} \mathbb{E}_{\bx_0 \sim p_0(\bx)}\mathbb{E}_{\bx|\bx_{0} \sim p_{0t}(\bx | \bx_0)} \left[\left\|\bS_t^{(\alpha)}(\bx; \theta) - \bS^{(\alpha)}_t(\bx | \bx_0) \right\|^2\right].
\end{align}
where $\bS^{(\alpha)}\left(\bx|\bx_0\right) = \frac{\Delta^{\frac{\alpha-2}{\alpha}}p_{0t}\left(\bx|\bx_0\right)}{p_{0t}\left(\bx|\bx_0\right)}$ is the conditional score function and $p_{0t}\left(\bx|\bx_0\right)$ is the conditional distribution given the starting point $\bx_{0}$ at $t=0$.
Conditioned on the starting point $\bx_{0}$, $p_{0t}\left(\bx | \bx_{0}\right)$ can be analytically obtained for common SDEs, such as L\'evy processes and 
the widely-used Ornstein–Uhlenbeck (OU) processes in image generation \cite{yoon2023scorebased}.
Specifically, these stochastic processes typically exhibit $\alpha$-stable distributions as their conditional distributions, with their conditional fractional score functions being linear. 
After obtaining the fractional score, we solve 
the LL-PDE \eqref{eq:LL_levy_pde} with the standard PINN.

\subsection{Score Fractional PINN (Score-fPINN)}
As FSM works only when conditional distributions are known,  we propose the Score-fPINN to solve general SDEs regardless of the conditional distributions.
Specifically, the Score-fPINN initially employs the Sliced Score Matching (SSM) to obtain the vanilla score function of the $p_t(\bx)$. 
Subsequently, this vanilla score function is integrated into the FPL equation, enabling the derivation of the fractional score via the standard PINN approach.

Observing that $\bS_t^{(2)}(\bx)=\nabla q=\nabla \log p_t(\bx)$, we obtain from the LL-PDE \eqref{eq:LL_levy_pde} that
\begin{align}
\partial_t \bS_t^{(2)}(\bx) = \nabla_{\bx}\left[\frac{1}{2}\nabla_{\bx} \cdot(\bG\bG^\mathrm{T}\bS_t^{(2)}) + \frac{1}{2}\Vert\bG^\mathrm{T}\bS_t^{(2)}\Vert^2 - \langle \boldsymbol{A}^\alpha, \bS_t^{(2)}\rangle - \nabla_{\bx} \cdot \boldsymbol{A}^\alpha\right] := \mathcal{L}_{\text{Score-fPDE}}\left[\bS_t^{(2)}, \bS_t^{(\alpha)}\right],
\end{align}
where $\bA^\alpha$ is defined in equation \eqref{eq:scorefpinn_A_alpha}.
We call this PDE Score-fPDE, where $\bdf$, $\bG$, and $\sigma$ are known.  The operator $\mathcal{L}_{\text{Score-fPDE}}$ takes the two scores as input variables. Thus, once we know $\bS^{(2)}_t(\bx)$, then $\bS^{\alpha}_t(\bx)$ inside $\bA^\alpha$ can be solved by learning from the PDE above since it is the only unknown in the PDE. Fortunately, the vanilla score $\bS^{(2)}_t(\bx)$ can be readily learned using a well-developed technique \cite{song2019sliced}, namely Sliced Score Matching (SSM).
Concretely, the SSM loss function to obtain a vanilla score function model $\bS^{(2)}_t\left(\bx;\theta^{(2)}\right)$ parameterized by $\theta^{(2)}$ such that $\bS^{(2)}_t\left(\bx;\theta^{(2)}\right) \approx \bS^{(2)}_t(\bx)$ is given by:
\begin{equation}\label{eq:loss_ssm_fractional}
\theta^{(2)} = \arg\min_{\theta^{(2)}} \mathbb{E}_{t \sim \text{Unif}[0,T]} \mathbb{E}_{\bx \sim p_t(\bx)}\left[\frac{1}{2}\left\| \bS_{t}^{(2)}\left(\bx;\theta^{(2)}\right) \right\|^2 + \nabla_{\bx}\cdot\bS_{t}^{(2)}\left(\bx;\theta^{(2)}\right)\right].
\end{equation} 
Note that SSM does not assume anything about the SDE type. SSM only needs SDE samples from $p_t(\bx)$, which can be obtained from any numerical SDE discretization scheme \cite{tretyakov2013fundamental,chen2023approximation}.

After obtaining the vanilla score function $\bS_{t}^{(2)}\left(\bx;\theta^{(2)}\right) \approx \bS^{(2)}_t(\bx)$ via SSM. We solve for the fractional score $\bS^{(\alpha)}_t(\bx)$ through score-fPINN on the score-fPDE:
\begin{equation}\label{eq:loss_fpinn}
L_{\text{Score-fPINN}}(\theta) = \mathbb{E}_{t \sim \text{Unif}[0, T]} \mathbb{E}_{\bx \sim p_t(\bx)} \left[\left(\partial_t \bS_t^{(2)}\left(\bx;\theta^{(2)}\right) - \mathcal{L}_{\text{Score-fPDE}}\left[\bS_t^{(2)}\left(\bx;\theta^{(2)}\right), \bS_t^{(\alpha)}\left(\bx;\theta\right)\right]\right)^2\right].
\end{equation}
After obtaining the fractional
score, we solve the equation \eqref{eq:LL_levy_pde} via the standard PINN. 


\begin{algorithm}
\caption{Fractional Score-based SDE solver.}
\begin{algorithmic}[1]
\State Obtain the approximated fractional score function $\bS_t^{(\alpha)}(\bx;\theta) \approx \bS_t^{(\alpha)}(\bx)$ via one of the two approaches:
\begin{itemize}
\item Fractional score matching (FSM): $\theta = \arg\min_\theta L^\alpha_{\text{Cond-SM}}(\theta)$ in equation (\ref{eq:loss_fsm}) if the SDE conditional distribution is tractable.
\item Score-fPINN: First obtain the vanilla score function $\bS^{(2)}_t\left(\bx;\theta^{(2)}\right) \approx \bS^{(2)}_t(\bx)$ from SSM loss in equation (\ref{eq:loss_ssm_fractional}). Then, optimize score-fPINN $\theta = \arg\min_\theta L_{\text{Score-fPINN}}(\theta)$ in equation (\ref{eq:loss_fpinn}).
\end{itemize}
\State Parameterize the LL model $q_t(\bx;\phi)$.
\State Obtain the LL by solving the LL-PDE \eqref{eq:LL_levy_pde} with $\bS^{(\alpha)}_t(\bx;\theta)$ as the score function, $\phi = \arg\min_\phi L_{\text{LL-PDE}}(\phi)$, where
\begin{equation}\label{eq:loss_ll_fpinn}
\begin{aligned}
&L_{\text{LL-PDE}}(\phi) = \lambda_{\text{initial}} \cdot \mathbb{E}_{\bx \sim p_0(\bx)} \left[\left(q_0(\bx;\phi) -\log p_0(\bx)\right)^2\right] + \\
&\quad\lambda_{\text{residual}} \cdot \mathbb{E}_{t \sim \text{Unif}[0, T]} \mathbb{E}_{\bx \sim p_t(\bx)} \left[\left(\partial_t q_t(\bx;\phi) - \mathcal{L}_{\text{LL-PDE}}\left[q_t(\bx;\phi), \bS_t^{(\alpha)}(\bx;\theta)\right]\right)^2\right].
\end{aligned}
\end{equation}
\end{algorithmic}
\label{algo:1}
\end{algorithm}

\subsection{Methods Summary and Comparison}\label{sec:comparison_fsm_scorefpinn}
We have introduced fractional score matching (FSM) or score-fPINN to obtain the fractional score function.
Then, we can infer LL by solving the LL PDE in equation \eqref{eq:LL_levy_pde} equipped with the fractional score. Similar to Score-PINN for FP equation \cite{hu2024score}, our fractional score-based SDE/FPL equation solver contains two stages. 
Algorithm \ref{algo:1} summarizes the above methodology.
Following Score-PINN \cite{hu2024score}, we parameterize the score and LL models differently.

In comparison, these two methods' second steps are the same. After obtaining the fractional score by either FSM or score-fPINN in the first step, we plug it into the FPL equation and transformed FPL into a second-order PDE.
Hence, FSM's and score-fPINN's computational costs are the same in the second stage, while only their first stages differ.

Regarding their first stages, FSM is more concise and efficient as its computational and loss function in equation \eqref{eq:loss_fsm} only requires the fractional score function model inference. However, it requires the known conditional distributions modeled by the SDE.
In contrast, Score-fPINN does not require such conditional distributions. But 
it employs PINN loss in equation (\ref{eq:loss_fpinn}) and thus necessitates computing derivatives of the neural network model with respect to the input, which results in higher computational cost.
Besides, the SSM loss in equation (\ref{eq:loss_ssm_fractional}) to obtain the vanilla score function also requires the first-order derivative of the score function model.
Thus, score-fPINN is much more expensive by design while score-fPINN applies to a broader range of SDE types. 

\subsection{Technical Details}
Due to the heavy-tail of $\alpha$-stable distribution as shown in Figure \ref{fig:compare_gaussian_levy}, the FPL equation with L\'evy noise and fractional Laplacian poses more challenges than the classical FP. 

First, the fractional score of a common distribution is usually unknown due to the computationally costly fractional Laplacian. 
When $\alpha=2$, the vanilla score function of the initial distribution $\bS^{(2)}_{t=0}(\bx)$ is usually known and can serve as the hard constraint to regularize the vanilla score function PINN model.
When $\alpha <2$, the fractional score requires numerical computations and thus introduces extra errors. 

Second, we use \textbf{smooth $L_1$ loss} instead of $L_2$ loss to robustify the optimization with extreme values/rare events produced by L\'evy noise following Yoon et al. \cite{yoon2023scorebased}.
Concretely, the smooth $L_1$ loss has a hyperparameter $\beta$ whose form is:
\begin{align}
\ell(\hat{y}, y) = 
\begin{cases} 
|\hat{y} - y|^2 & \text{if } |\hat{y} - y| < \beta \\
2\beta|\hat{y} - y|-\beta^2 & \text{if } |\hat{y} - y| \geq \beta
\end{cases}
\end{align}
Here, $\hat{y}$ denotes the model prediction, and $y$ is the ground truth label.
When $\beta \rightarrow \infty$, it converges to $L_2$ loss. 
When the prediction and label diverge significantly ($|\hat{y} - y| \geq \beta$), such as when rare events or extreme values occur in the L\'evy process, employing an $L_1$ loss generates gradients with a smaller scale, ensuring stability. Conversely, when the prediction and label do not differ significantly ($|\hat{y} - y| < \beta$), a standard $L_2$ loss is used.

Last, we may match both $\bS^{(2)}_t(\bx)$ and $\bS^{(\alpha)}_t(\bx)$ like Score-PINN \cite{hu2024score} to transform the FPL equation into ODE further. However, learning two score functions and plugging them into the FPL equation will lead to substantial error.
In practice, the inherent error introduced by inserting two score functions is significant to the extent that PINN cannot solve for the solution of the corresponding LL-ODE. 
Conversely, matching only one fractional score function proves accurate enough, enabling PINN to extract the likelihood from the corresponding LL-PDE.


\subsection{Comparison with Related Work by Yoon et al.}
We compare our fractional score-based SDE solver with the generative model based on the fractional score by Yoon et al. \cite{yoon2023scorebased}. 
We employ the fractional score to solve the PDF and LL modeling by the SDE and FPL equations. 
On the other hand, the generative model generates data through the SDE without explicitly focusing on the PDF and LL, emphasizing the generated data's quality. Furthermore, when solving the SDE, we already know the SDE's initial conditions and form, particularly the drift and diffusion coefficients, which may be nonlinear or more complex. In contrast, in the generative model, the initial conditions are unknown data distributions. The generative model utilizes the SDE to transform the data distribution into pure L\'evy noise and then reverses the SDE. The reverse SDE serves as a pathway from noise to images, enabling image generation, with its drift coefficient associated with the fractional score function. Hence, the generative model matches the fractional score to invert the SDE and generate data. 
Finally, the generative model employs relatively simple SDEs, such as Ornstein-Uhlenbeck (OU) processes and L\'evy processes, to generate images since these processes provable convert any unknown data distribution to pure L\'evy noises, while we can address more general SDEs. The comparisons and differing methodologies above also apply to the distinction between our SDE/FPL equation solver and other score-based generative models \cite{song2021scorebased}.

\section{Computational Experiment}

In this section, we conduct numerical experiments to evaluate the performance of our fractional score-based SDE solver under various experimental settings, including anisotropic Ornstein-Uhlenbeck (OU) processes and stochastic processes with dual noise components of Brownian and L\'evy, ultimately testing SDE with nonlinear drift. 
We test analytically solvable SDEs or SDEs with a special structure, which we can solve accurately.
We further manipulate various initial distributions to test the fractional score-matching approach. We also validate the effectiveness of details designed in the methodology, such as the smooth $L_1$ loss.


We take $\alpha \in \{1.95,1.85,1.75\}$, dimension $d$ in $\{10,20,50,100\}$.
We will investigate the effect of dimension and $\alpha$ on our methods' performances. 
The score models in FSM and score-fPINN are all four-layer fully connected networks whose input and output dims are the same as the SDE dimensionality $d$, while the hidden dimension is 128 for all cases. 
The LL model is also a 4-layer fully connected network whose input dimension is $d$, output dimension is 1, and hidden dimension is 128 for all cases. 
We adopt the following model structure to satisfy the initial condition with hard constraint and to avoid the boundary/initial loss \cite{lu2021physics} for the LL model
$q_t(\bx;\phi) = \text{NN}(\bx, t; \phi) t -\log p_0(\bx).
$
and 
$
\bS^{(2)}_t\left(\bx;\theta^{(2)}\right) = \text{NN}(\bx, t; \theta^{(2)}) t - \nabla \log p_0(\bx)
$. 


In the smooth $L_1$ loss, we 
 take $\beta=1$. In Section \ref{ssec:example1-simple}, we also test the influence of different choices of $\beta$. 
The score and LL models are trained via Adam \cite{kingma2014adam} for 100K and 10K epochs, respectively, with an initial learning rate of 1e-3, which exponentially decays with a decay rate of 0.9 for every 10K epochs. 
We select 10K random residual points along the SDE trajectory with uniform time steps $t$ at each Adam epoch for all methods in training score and LL. 
We randomly sample 10K fixed testing points along the SDE trajectory. 
For the test points, we delete test points of rare points far away from $\mathbf{0}$, see Figure \ref{fig:compare_gaussian_levy} for examples, which is mathematically defined as those with the lowest 10\% LL.

All experiments are conducted on an NVIDIA A100 GPU with 80GB of memory. Due to its efficient automatic differentiation, we implement our algorithm using JAX \cite{jax2018github}. All experiments are computed five times with five independent random seeds for reproducibility.

\subsection{High-Dimensional Anisotropic Basic SDE}\label{ssec:example1-simple}
\subsubsection{SDE Formulation}
We consider a basic SDE with an anisotropic multivariate Gaussian as the initial condition:
\begin{align}\label{eq:basic-sde}
d\bx = d\bw_t + d\bL_t^\alpha,  \quad p_0(\bx) \sim \mathcal{N}(0, \bSigma),
\end{align}
where $\bSigma =\operatorname{diag}(\lambda_1, \lambda_2, \cdots, \lambda_d)$ where $\lambda_{2i} \sim \text{Unif}[1, 2]$ and $\lambda_{2i+1} = 1/\lambda_{2i}$ are randomly generated for an anisotropic initial condition and SDE.
This SDE gradually injects Gaussian and L\'evy noises into the initial condition, and its exact solution is the  sum of  a Gaussian $\mathcal{N}(0, \bSigma + t\bI)$ and the L\'evy noise $\bL_t^\alpha$.
Despite its simple form, this SDE is still anisotropic due to the initial condition and effectively high-dimensional due to the inseparable L\'evy noise, i.e., this SDE cannot be simplified to low-dimensional cases. Thus, traditional methods cannot solve it.

\subsubsection{Hyperparamter Setting}
Here we set the terminal time $T=1$.
The networks are $q_t(\bx;\phi) = \text{NN}(\bx, t; \phi) t -\frac{d}{2}\log(2\pi) - \frac{1}{2}\bx^\mathrm{T}\bSigma^{-1}\bx,
$ and $\bS^{(2)}_t\left(\bx;\theta^{(2)}\right) = \text{NN}(\bx, t; \theta^{(2)}) t - \bSigma^{-1}\bx$, where both networks are fully-connected neural networks as described at the beginning of the section.
The exact distribution is the one for the sum of a Gaussian $\mathcal{N}(0, \bSigma + t\bI)$ and the L\'evy noise $\bL_t^\alpha$. The distribution is computed via Monte Carlo simulation with $10^7$ samples. 
The evaluation criterion is the relative $L_2$ error of the LL.

\subsubsection{Main Results}

\begin{table}[htbp]
\centering
\begin{tabular}{|c|c|c|c|c|c|c|c|c|c|}
\hline
Method & Dimension & alpha & Time & $L_2$ & Method & Dimension & alpha & Time & $L_2$ \\ \hline
\multirow{12}{*}{FSM} & \multirow{3}{*}{100} & 1.75 & \multirow{3}{*}{12min} & 2.81E-2 & \multirow{12}{*}{Score-fPINN} & \multirow{3}{*}{100} & 1.75 & \multirow{3}{*}{59min} & 2.83E-2 \\ \cline{3-3} \cline{5-5} \cline{8-8} \cline{10-10} 
 &  & 1.85 &  & 2.48E-2 &  &  & 1.85 &  & 2.70E-2 \\ \cline{3-3} \cline{5-5} \cline{8-8} \cline{10-10} 
 &  & 1.95 &  & 2.48E-2 &  &  & 1.95 &  & 2.47E-2 \\ \cline{2-5} \cline{7-10} 
 & \multirow{3}{*}{50} & 1.75 & \multirow{3}{*}{11min} & 2.16E-2 &  & \multirow{3}{*}{50} & 1.75 & \multirow{3}{*}{56min} & 2.17E-2 \\ \cline{3-3} \cline{5-5} \cline{8-8} \cline{10-10} 
 &  & 1.85 &  & 1.94E-2 &  &  & 1.85 &  & 1.98E-2 \\ \cline{3-3} \cline{5-5} \cline{8-8} \cline{10-10} 
 &  & 1.95 &  & 1.52E-2 &  &  & 1.95 &  & 1.44E-2 \\ \cline{2-5} \cline{7-10} 
 & \multirow{3}{*}{20} & 1.75 & \multirow{3}{*}{7min} & 1.23E-2 &  & \multirow{3}{*}{20} & 1.75 & \multirow{3}{*}{50min} & 1.28E-2 \\ \cline{3-3} \cline{5-5} \cline{8-8} \cline{10-10} 
 &  & 1.85 &  & 1.02E-2 &  &  & 1.85 &  & 1.09E-2 \\ \cline{3-3} \cline{5-5} \cline{8-8} \cline{10-10} 
 &  & 1.95 &  & 1.02E-2 &  &  & 1.95 &  & 1.08E-2 \\ \cline{2-5} \cline{7-10} 
 & \multirow{3}{*}{10} & 1.75 & \multirow{3}{*}{4min} & 6.59E-3 &  & \multirow{3}{*}{10} & 1.75 & \multirow{3}{*}{42min} & 6.72E-3 \\ \cline{3-3} \cline{5-5} \cline{8-8} \cline{10-10} 
 &  & 1.85 &  & 6.21E-3 &  &  & 1.85 &  & 6.19E-3 \\ \cline{3-3} \cline{5-5} \cline{8-8} \cline{10-10} 
 &  & 1.95 &  & 4.43E-3 &  &  & 1.95 &  & 4.58E-3 \\ \hline
\end{tabular}
\caption{Main results for the basic SDE \eqref{eq:basic-sde} up to 100D with various $\alpha$. We present the relative $L_2$ errors of FSM and Score-fPINN, which perform similarly. We also present the total running time for both algorithms to demonstrate their scalability.}
\label{tab:score_fpinn_basic_sde}
\end{table}

The main results are shown in Table \ref{tab:score_fpinn_basic_sde}.
FSM and score-fPINN can achieve low errors ($\sim 1\%$) in high dimensions up to 100D.
The two methods' total running time grows sublinearly across different dimensions thanks to the grid-less and mesh-free PINN and score-matching training
But score-fPINN is more expensive than FSM, as discussed in Section \ref{sec:comparison_fsm_scorefpinn}. 
As dimension increases, the PDF/LL value becomes smaller, decaying exponentially, and tends to be more unstable, and thus, the prediction error becomes larger,
As $\alpha$ decreases, the $\alpha$-stable distribution's support becomes wider, and there are more rare events and extreme values, posing an additional challenge to PINN.
These are all reflected in the results.
Overall, the main results demonstrate the stable performance and scalability of our proposed fractional score-based SDE/FPL equation solver. 

\subsubsection{Additional Study: Effect of Smooth $L_1$ Loss}

\begin{table}[htbp]
\centering
\begin{tabular}{|c|c|c|c|c|}
\hline
Method & Dimension & alpha & Beta & $L_2$ \\ \hline
\multirow{5}{*}{FSM} & 100 & 1.75 & 0.01 & 3.08E-2 \\ \cline{2-5} 
 & 100 & 1.75 & 0.1 & 2.88E-2 \\ \cline{2-5} 
 & 100 & 1.75 & 1 & \textbf{2.81E-2} \\ \cline{2-5} 
 & 100 & 1.75 & 10 & 3.30E-2 \\ \cline{2-5} 
 & 100 & 1.75 & 100 & 7.20E-2 \\ \hline
\end{tabular}
\caption{Results for the effect of $\beta$ in the smooth $L_1$ loss. The best performance is achieved with $\beta = 1$, which is bolded.}
\label{tab:score_fpinn_effect_l1_loss}
\end{table}

This additional study investigates the effect of $\beta$ in the smooth $L_1$ loss. We kept all the same hyperparameters as in the main results, except we manipulated $\beta$ in the smooth $L_1$ loss in $\{0.01, 0.1, 1, 10, 100\}$. We only study the most difficult 100D case with $\alpha = 1.75$ and test the FSM method since the two methods perform similarly.

The results are shown in Table \ref{tab:score_fpinn_effect_l1_loss}. With increasing $\beta$ in the smooth $L_1$ loss in $\{1, 10, 100\}$, we notice a drop in the final relative $L_2$ error performance on LL. This illustrates the importance of the smooth $L_1$ loss on the robustness of optimization in the presence of extreme values and rare events produced by stable distributions. With larger $\beta$, the smooth $L_1$ loss will gradually converge to the $L_2$ loss and become more sensitive to extreme values, thus deteriorating the performance.
On the other hand, if we choose a small $\beta$, e.g., $\beta = 0.01$ and $\beta = 0.1$, its optimization effect will deteriorate too due to the following disadvantages of $L_1$ loss.
The gradient of the $L_1$ loss being constant (except at zero, where it is undefined) can indeed impact the convergence speed of the optimization process. 
Since the gradient does not scale with the error, significant errors are penalized no more than small errors in gradients. This can lead to inefficient convergence in scenarios where the errors vary significantly in magnitude.
When the prediction exactly matches the target (i.e., error = 0), the derivative of the $L_1$ loss is undefined, which can cause issues in gradient-based optimization methods. 
The non-differentiability at zero can make the optimization landscape challenging, especially for optimization algorithms that rely heavily on the smoothness of the function, such as some variants of gradient descent.

In conclusion, this additional demonstrates the effectiveness and importance of employing smooth $L_1$ loss and underscores the importance of choosing an appropriate $\beta$. The se;ection guidelines and underlying mechanisms are also explained.

\subsection{High-Dimensional Anisotropic SDE with Complicated Diffusion Coefficient}
\subsubsection{SDE Formulation}
We consider the anisotropic SDE with Brownian and L\'evy noises and unit Gaussian as the initial condition:
\begin{align}\label{eq:sde-complicated}
d\bx = (\bB + t \bI) d\bw_t + d\bL_t^\alpha, \quad p_0(\bx) \sim \mathcal{N}(0, \bI).
\end{align}
The exact solution is the  sum of  a Gaussian $p_t(\bx) \sim \mathcal{N}(0,  \boldsymbol{\Sigma}_t)$ and the L\'evy noise $\bL_t^\alpha$, where
\begin{align}
\boldsymbol{\Sigma}_t &= \bI + \int_0^t \left[ (\bB +s \bI) (\bB +s \bI)^\mathrm{T}\right]ds  = \left(1 + \frac{t^3}{3}\right)\bI + t\bB\bB^\mathrm{T} + \frac{t^2}{2}(\bB + \bB^\mathrm{T}).
\end{align}
Here, we generate $\bB = \boldsymbol{Q}\boldsymbol{\Gamma}$ where $\boldsymbol{Q}$ is a random orthogonal matrix and $\boldsymbol{\Gamma}$ is a diagonal matrix generated in the same way used in the previous example of basic SDE. 
The anisotropic Gaussian $\mathcal{N}(0, \boldsymbol{\Sigma}_t)$ covariance matrix's eigenspace evolves if $\bB$ is not orthogonal and not symmetric, posing an additional challenge. Compared to the previous case study, we attempt to test more difficult and effective high-dimensional SDE with complicated diffusion coefficient $\bG(\bx, t) = \bB + t \bI$. This SDE is anisotropic due to the diffusion coefficient and effectively high-dimensional due to the inseparable L\'evy noise, i.e., this SDE cannot be simplified to low-dimensional cases.

\subsubsection{Hyperparamter Setting}
Here we set the terminal time $T=1$ and present the relative $L_2$ errors of the LL at $T=1$. 
%
%
%
%
The reference, i.e., the LL of the distribution that is the sum of a Gaussian $\mathcal{N}(0, \bSigma_t)$ and the L\'evy noise $\bL_t^\alpha$, is computed via Monte Carlo simulation with $10^7$ samples. Unlike in Section \ref{ssec:example1-simple}, sampling along the SDE trajectory requires the heavy computation of $\bSigma_t^{1/2}$. 
Hence, we do not resample $t$ at each training iteration and precompute $\bSigma_t^{1/2}$. Specifically, after fixing $t$, we compute $\bSigma_t^{1/2}$ for each $t$ as follows.
We conduct eigendecomposition for $\bSigma_t = \boldsymbol{Q}_t^\mathrm{T} \boldsymbol{\Gamma}_t \boldsymbol{Q}_t$ where $\boldsymbol{Q}_t$ is orthogonal and $\boldsymbol{\Gamma}_t$ is diagonal. Then, $\bSigma_t^{1/2}$ can be computed via $\bSigma_t^{1/2} = \boldsymbol{Q}_t^\mathrm{T} \boldsymbol{\Gamma}_t^{1/2} \boldsymbol{Q}_t$.
%

\subsubsection{Main Results}

\begin{table}[htbp]
\centering
\begin{tabular}{|c|c|c|c|c|c|c|c|c|c|}
\hline
Method & Dimension & alpha & Time & $L_2$ & Method & Dimension & alpha & Time & $L_2$ \\ \hline
\multirow{12}{*}{FSM} & \multirow{3}{*}{100} & 1.75 & \multirow{3}{*}{25min} & 4.98E-2 & \multirow{12}{*}{Score-fPINN} & \multirow{3}{*}{100} & 1.75 & \multirow{3}{*}{72min} & 5.00E-2 \\ \cline{3-3} \cline{5-5} \cline{8-8} \cline{10-10} 
 &  & 1.85 &  & 2.95E-2 &  &  & 1.85 &  & 3.02E-2 \\ \cline{3-3} \cline{5-5} \cline{8-8} \cline{10-10} 
 &  & 1.95 &  & 2.74E-2 &  &  & 1.95 &  & 2.77E-2 \\ \cline{2-5} \cline{7-10} 
 & \multirow{3}{*}{50} & 1.75 & \multirow{3}{*}{19min} & 3.79E-2 &  & \multirow{3}{*}{50} & 1.75 & \multirow{3}{*}{68min} & 3.78E-2 \\ \cline{3-3} \cline{5-5} \cline{8-8} \cline{10-10} 
 &  & 1.85 &  & 2.56E-2 &  &  & 1.85 &  & 2.58E-2 \\ \cline{3-3} \cline{5-5} \cline{8-8} \cline{10-10} 
 &  & 1.95 &  & 1.79E-2 &  &  & 1.95 &  & 1.75E-2 \\ \cline{2-5} \cline{7-10} 
 & \multirow{3}{*}{20} & 1.75 & \multirow{3}{*}{14min} & 1.92E-2 &  & \multirow{3}{*}{20} & 1.75 & \multirow{3}{*}{62min} & 1.98E-2 \\ \cline{3-3} \cline{5-5} \cline{8-8} \cline{10-10} 
 &  & 1.85 &  & 1.28E-2 &  &  & 1.85 &  & 1.30E-2 \\ \cline{3-3} \cline{5-5} \cline{8-8} \cline{10-10} 
 &  & 1.95 &  & 1.14E-2 &  &  & 1.95 &  & 1.23E-2 \\ \cline{2-5} \cline{7-10} 
 & \multirow{3}{*}{10} & 1.75 & \multirow{3}{*}{12min} & 1.32E-2 &  & \multirow{3}{*}{10} & 1.75 & \multirow{3}{*}{57min} & 1.11E-2 \\ \cline{3-3} \cline{5-5} \cline{8-8} \cline{10-10} 
 &  & 1.85 &  & 9.41E-3 &  &  & 1.85 &  & 8.99E-3 \\ \cline{3-3} \cline{5-5} \cline{8-8} \cline{10-10} 
 &  & 1.95 &  & 5.21E-3 &  &  & 1.95 &  & 5.62E-3 \\ \hline
\end{tabular}
\caption{Main results for the SDE with a complicated diffusion coefficient \eqref{eq:sde-complicated} up to 100D with various $\alpha$. We present the relative $L_2$ errors of FSM and Score-fPINN, which perform similarly. We also present the total running time for both algorithms to demonstrate  scalability.}
\label{tab:score_fpinn_complicated_diffusion_sde}
\end{table}

The main results are shown in Table \ref{tab:score_fpinn_complicated_diffusion_sde}.
FSM and score-fPINN can achieve low errors ($\sim 2\%$) in high dimensions up to 100D. 
The errors are larger than the previous basic SDE due to the complicated diffusion coefficient in the current case, causing the Gaussian component in the exact solution to have varying eigenspace with time evolution.
The two methods' total running time grows sublinearly across different dimensions as before but they are slightly more expensive than in the previous basic SDE \eqref{eq:basic-sde} since the computation of the diffusion coefficient takes extra time.
 We observe that score-fPINN is more expensive than FSM, as discussed in Section \ref{sec:comparison_fsm_scorefpinn}. 
Like the previous basic SDE results, the relative $L_2$ error of both methods are similar and grow slightly with the increasing SDE dimension and decreasing $\alpha$.
Overall, the main results demonstrate the stable performance and scalability of our proposed fractional score-based SDE/FPL equation solver on SDE with a complicated diffusion coefficient.

\subsubsection{Additional Study: Effect of Laplacian Initial Distribution}

\begin{table}[htbp]
\centering
\begin{tabular}{|c|c|c|c|c|c|c|c|c|c|}
\hline
Method & Dimension & alpha & Time & $L_2$ & Method & Dimension & alpha & Time & $L_2$ \\ \hline
\multirow{3}{*}{FSM} & \multirow{3}{*}{100} & 1.75 & \multirow{3}{*}{95min} & 6.57E-2 & \multirow{3}{*}{Score-fPINN} & \multirow{3}{*}{100} & 1.75 & \multirow{3}{*}{297min} & 6.68E-2 \\ \cline{3-3} \cline{5-5} \cline{8-8} \cline{10-10} 
 &  & 1.85 &  & 6.50E-2 &  &  & 1.85 &  & 6.42E-2 \\ \cline{3-3} \cline{5-5} \cline{8-8} \cline{10-10} 
 &  & 1.95 &  & 6.28E-2 &  &  & 1.95 &  & 6.19E-2 \\ \hline
\end{tabular}
\caption{Additional results for the SDE with a complicated diffusion coefficient \eqref{eq:sde-complicated} up to 100D with various $\alpha$, whose initial distribution is changed to be Laplacian for more SDE problem complexity. We present the relative $L_2$ errors of FSM and Score-fPINN, which perform similarly. We also present the total running time for both algorithms to demonstrate scalability.}
\label{tab:score_fpinn_complicated_diffusion_sde_laplace}
\end{table}

This additional study investigates the effect of changing the initial distribution to be a unit Laplacian distribution, which causes the exact solution to be the sum of the Gaussian $\mathcal{N}(0, \bSigma_t - \bI)$, the unit Laplacian, and the L\'evy noise $\bL_t^\alpha$. The mixed exact solution poses more challenges to the SDE solver. We kept the hyperparameters the same as in the main results, with the only difference being that we will use boundary loss with 20 weight for the LL model instead of the hard constraint due to the non-smoothness of the Laplacian PDF and LL. Also, we extend the training of the LL model from the previous 10K epochs to 100K epochs. This is because we now require the boundary loss for the non-smooth Laplacian initial condition instead of the hard constraint, which requires more epochs for the LL model to optimize the additional loss function. We test both FSM and score-fPINN in 100D.

The additional results are shown in Table \ref{tab:score_fpinn_complicated_diffusion_sde_laplace}. 
Overall, these additional results demonstrate the stable performance and scalability of our proposed score-fPINN on an SDE with a complicated diffusion coefficient plus the Laplacian initial condition, whose exact solution is a mixture of anisotropic Gaussian, Laplacian, and stable distributions.

\subsection{High-Dimensional Anisotropic OU Processes with Linear Drift}
\subsubsection{SDE Formulation}
We consider the OU process and an anisotropic multivariate Gaussian as the initial condition:
\begin{align}\label{eq:sde-linear-nobrown}
d\bx = -\frac{\bx}{\alpha} dt + d\bL_t^\alpha,  \quad p_0(\bx) \sim \mathcal{N}(0, \bSigma),
\end{align}
where $\bSigma =\operatorname{diag}(\lambda_1, \lambda_2, \cdots, \lambda_d)$ where $\lambda_{2i} \sim \text{Unif}[1, 2]$ and $\lambda_{2i+1} = 1 / \lambda_{2i}$.
This OU process is interesting since it gradually transforms any initial distribution to the unit $\alpha$-stable distribution as $t \rightarrow \infty$. The score-based generative model \cite{yoon2023scorebased} utilizes this OU process to transform the data distribution into pure L\'evy noise and then reverses the SDE. The reverse SDE serves as a pathway from noise to images, enabling image generation, with its drift coefficient associated with the fractional score function. Mathematically, the exact solution to this OU process is given by
\begin{align}
\bx_t = \exp\left(-\frac{t}{\alpha}\right)\bx_0 + \mathcal{S}\alpha\mathcal{S}^d\left(\left(1 - \exp\left(-t\right)\right)^{1/\alpha}\right), \quad \text{ in distribution}.
\end{align}
The exact solution is the weighted sum of the initial condition and an $\alpha$-stable process, which can be computed via Monte Carlo simulation. This SDE is still anisotropic due to the initial condition and effectively high-dimensional due to the inseparable L\'evy noise, i.e., this SDE cannot be simplified to low-dimensional cases. 

\subsubsection{Hyperparamter Setting}
Here the terminal time $T=0.5$. 
%
%
%
We adopt the following models:
$
q_t(\bx;\phi) = \text{NN}(\bx, t; \phi) t -\frac{d}{2}\log(2\pi) - \frac{1}{2}\bx^\mathrm{T}\bSigma^{-1}\bx,
$
where $\text{NN}(\bx, t; \phi)$ is the fully connected neural network and 
$
\bS^{(2)}_t\left(\bx;\theta^{(2)}\right) = \text{NN}(\bx, t; \theta^{(2)}) t - \bSigma^{-1}\bx,
$
where $\text{NN}(\bx, t; \theta^{(2)})$ is the fully connected neural network. Both networks are described at the beginning of the section.

The reference, i.e., the LL of the distribution, the sum of a Gaussian and the L\'evy noise, is computed via Monte Carlo simulation with $10^7$ samples. 
The evaluation criterion is the relative $L_2$ error of the LL.

\begin{table}[htbp]
\begin{tabular}{|c|c|c|c|l|c|c|c|c|l|}
\hline
Method & Dimension & alpha & Time & \multicolumn{1}{c|}{$L_2$} & Method & Dimension & alpha & Time & \multicolumn{1}{c|}{$L_2$} \\ \hline
\multirow{12}{*}{FSM} & \multirow{3}{*}{100} & 1.75 & \multirow{3}{*}{21min} & 3.98E-2 & \multirow{12}{*}{Score-fPINN} & \multirow{3}{*}{100} & 1.75 & \multirow{3}{*}{68min} & \multicolumn{1}{c|}{3.68E-2} \\ \cline{3-3} \cline{5-5} \cline{8-8} \cline{10-10} 
 &  & 1.85 &  & 3.90E-2 &  &  & 1.85 &  & 3.57E-2 \\ \cline{3-3} \cline{5-5} \cline{8-8} \cline{10-10} 
 &  & 1.95 &  & 3.87E-2 &  &  & 1.95 &  & 3.44E-2 \\ \cline{2-5} \cline{7-10} 
 & \multirow{3}{*}{50} & 1.75 & \multirow{3}{*}{17min} & 1.23E-2 &  & \multirow{3}{*}{50} & 1.75 & \multirow{3}{*}{56min} & 1.34E-2 \\ \cline{3-3} \cline{5-5} \cline{8-8} \cline{10-10} 
 &  & 1.85 &  & 1.15E-2 &  &  & 1.85 &  & 1.20E-2 \\ \cline{3-3} \cline{5-5} \cline{8-8} \cline{10-10} 
 &  & 1.95 &  & 1.15E-2 &  &  & 1.95 &  & 1.18E-2 \\ \cline{2-5} \cline{7-10} 
 & \multirow{3}{*}{20} & 1.75 & \multirow{3}{*}{15min} & 4.67E-3 &  & \multirow{3}{*}{20} & 1.75 & \multirow{3}{*}{41min} & 5.33E-3 \\ \cline{3-3} \cline{5-5} \cline{8-8} \cline{10-10} 
 &  & 1.85 &  & 3.97E-3 &  &  & 1.85 &  & 4.08E-3 \\ \cline{3-3} \cline{5-5} \cline{8-8} \cline{10-10} 
 &  & 1.95 &  & 2.54E-3 &  &  & 1.95 &  & 3.28E-3 \\ \cline{2-5} \cline{7-10} 
 & \multirow{3}{*}{10} & 1.75 & \multicolumn{1}{l|}{\multirow{3}{*}{13min}} & 3.15E-3 &  & \multirow{3}{*}{10} & 1.75 & \multirow{3}{*}{32imn} & 4.88E-3 \\ \cline{3-3} \cline{5-5} \cline{8-8} \cline{10-10} 
 &  & 1.85 & \multicolumn{1}{l|}{} & 2.84E-3 &  &  & 1.85 &  & 4.34E-3 \\ \cline{3-3} \cline{5-5} \cline{8-8} \cline{10-10} 
 &  & 1.95 & \multicolumn{1}{l|}{} & 2.16E-3 &  &  & 1.95 &  & 3.04E-3 \\ \hline
\end{tabular}
\caption{Main results for the OU process up to 100D with various $\alpha$. We present the relative $L_2$ errors of FSM and Score-fPINN, which perform similarly. We also present the total running time for both algorithms to demonstrate scalability.}
\label{tab:score_fpinn_ou}
\end{table}

\subsubsection{Main Results}
The main results are shown in Table \ref{tab:score_fpinn_ou}.
FSM and score-fPINN can achieve low errors ($\sim 1\%$) in high dimensions up to 100D. 
The two methods' total running time grows sublinearly across different dimensions. 
As dimension increases, the PDF/LL value becomes smaller, decaying exponentially, and tends to be more unstable, and thus, the prediction error becomes larger. 
As $\alpha$ decreases, the $\alpha$-stable distribution's support becomes wider, and there are more rare events and extreme values, posing an additional challenge to PINN.
These results demonstrate our methods' capability to deal with anisotropic OU processes with linear drifts.

\subsection{High-Dimensional SDE with Complicated Drift Coefficient}

\subsubsection{SDE Formulation}
We have already tested several anisotropic SDEs with simple drift coefficients. 
Since obtaining a reference solution for general SDE with L\'evy noise and complicated drift is difficult, we design a high-dimensional FPL equation with a low effective dimension, which can be solved via traditional methods for reference.
An intuitive example is the simple L\'evy process $d\bx = d\bL^\alpha_t$ with the isotropic initial condition $p_0(\bx) \sim \mathcal{N}(0, \bI)$. Since the initial condition and the injected noise are all isotropic, the solution is only a function of $r = \Vert\bx\Vert$.
In the polar coordinate, it becomes a 1D problem. Garofalo \cite{garofalo2017fractional} gives the simplified fractional Laplacian in polar coordinate if the function is only about $r$. Thus, the intrinsically low-dimensional problem can be solved via traditional methods for score-fPINN's reference.
Notably, the high-dimensional $\alpha$-stable distribution/L\'evy noise is inseparable, though it is isotropic across different dimensions. 
More generally, the following high-dimensional FPL equation can be reduced to 1D problems:
$
d\bx =f \left( \Vert \bx \Vert\right)\bx dt + d\bL_t^\alpha
$
where $f$ can be any smooth function. Here we consider a nonlinear hyperbolic tangent drift:
\begin{align}\label{eq:sde-tanh-drift}
d\bx = -\bx \tanh\left(\Vert \bx \Vert / \sqrt{d}\right) dt + d\bL_t^\alpha. \quad p_0(\bx) \sim \mathcal{N}(0, \bI).
\end{align}

\subsubsection{Reference Generation}
Since the exact solution is isotropic and only concerns $r$ in the polar coordinate, we can generate reference accurately using kernel density estimation (KDE) thanks to its low-dimensional substructure.
We use the Euler-Maruyama scheme to discretize the SDE to get SDE samples/trajectories. 
The step size for the Euler-Maruyama scheme to discretize the SDE is $0.003$.
After that, we compute the norm of these samples and use KDE to estimate the distribution of the SDE samples' norms. Then, KDE provides the distribution of $r = \Vert \bx \Vert$ of the SDE solution. Since the exact solution is isotropic and only concerns $r$, we multiply the KDE results by the normalizing constant $(d-1)\log(r) - (d / 2 - 1) \log(2) - \log \Gamma(d / 2) + d /2 \log(2\pi)$ to obtain the final LL reference.

\subsubsection{Hyperparamter Setting}
We set $\alpha \in \{1.95,1.85,1.75\}$ and $d=100$ and present the relative $L_2$ errors of the LL at the terminal time $T=0.3$. 
%
%
%
%
%
%
%
In this example, we select 1K random residual points along the SDE trajectory at each Adam epoch in the training score and LL and 10K fixed testing points at the terminal time only. 
%
We adopt the following models:
$
q_t(\bx;\phi) = \text{NN}(\bx, t; \phi) t -\frac{d}{2}\log(2\pi) - \frac{1}{2}\bx^\mathrm{T}\bx,
$
where $\text{NN}(\bx, t; \phi)$ is the fully connected neural network and 
$
\bS^{(2)}_t\left(\bx;\theta^{(2)}\right) = \text{NN}(\bx, t; \theta^{(2)}) t - \bx,
$
where $\text{NN}(\bx, t; \theta^{(2)})$ is the fully connected neural network.
Both networks are described at the beginning of the section. 

\begin{table}[htbp]
\centering
\begin{tabular}{|c|c|c|c|c|}
\hline
Method & Dimension & alpha & Time & $L_2$ \\ \hline
\multirow{3}{*}{Score-fPINN} & \multirow{3}{*}{100} & 1.75 & \multirow{3}{*}{50min} & 8.51E-3 \\ \cline{3-3} \cline{5-5} 
 &  & 1.85 &  & 4.97E-3 \\ \cline{3-3} \cline{5-5} 
 &  & 1.95 &  & 2.39E-3 \\ \hline
\end{tabular}
\caption{Results for SDE \eqref{eq:sde-tanh-drift} with nonlinear drift coefficient in 100D with various $\alpha$.}
\label{tab:scorefpinn_nonlinear_drift}
\end{table}

\subsubsection{Main Results}
Table \ref{tab:scorefpinn_nonlinear_drift} shows the main results for SDE  \eqref{eq:sde-tanh-drift} with a complicated drift coefficient. Note that the conditional distribution of the PDF modeled by this SDE is unknown and thus FSM is not applicable. We only present results for score-fPINN.
Similar to previous experiments,  score-fPINN achieves stable performances within a reasonable running time. Its relative error remained stable for tested $\alpha$'s in  100D. This illustrates that score-fPINN is applicable across various SDE scenarios, including basic SDE and SDEs with complicated 
coefficients.

\section{Summary}
We considered computational methods for high-dimensional stochastic differential equations (SDEs) with L\'evy noise and their corresponding Fokker-Planck-L\'evy (FPL) equations.
We proposed a novel approach that accurately infers a log-likelihood (LL) solution by employing a fractional score-based SDE solver to fit the fractional score function. 
We demonstrated the numerical stability and the critical role of the fractional score function in modeling the SDE distribution. 
We introduced two fitting methods, Fractional Score Matching (FSM) and Score-fPINN, and thoroughly compare them in computational complexity and generality. 
Specifically, FSM is more concise and faster to train but only applies to SDEs with known conditional distributions. Score-fPINN, due to its need for computing model derivatives, is slower but more versatile. 
Efficient LL inference can be achieved after fitting the fractional score function. 
Our experiments confirm the stability and performance of the proposed SDE solver across various SDEs, demonstrating true lift of CoD. Importantly, our method outperforms traditional approaches and maintains stable computational costs as dimensions increase. 
The proposed method addresses challenges in high-dimensional stochastic systems, paving the way for further exploration and application in various scientific and engineering fields. Future work will include accelerating training by speeding up the sampling of stable distributions and L\'evy noises, which is a current bottleneck in training such models compared to the fast sampling of conventional Gaussian distributions.

\bibliographystyle{plain}
\bibliography{references}
\end{document}